\documentclass[11pt,a4paper]{article}
\usepackage[hyperref]{naaclhlt2018}
\usepackage{times}
\usepackage{latexsym}
\usepackage{amsmath}
\usepackage{siunitx}

\usepackage{url}

\aclfinalcopy 

\title{Pieces of Eight: 8-bit Neural Machine Translation\thanks{~~A piece of eight was a Spanish dollar that was divided into 8 reales, also known as Real de a Ocho.}}

\author{Jerry Quinn ~~~~~~~~~~Miguel Ballesteros\\
IBM Research, \\ 1101 Kitchawan Road, Route 134
Yorktown Heights, NY 10598. U.S \\
{ \sf jlquinn@us.ibm.com, miguel.ballesteros@ibm.com} }
\date{}

\begin{document}
\maketitle
\begin{abstract}
Neural machine translation has achieved levels of fluency and adequacy that would have been surprising a short time ago. Output quality is extremely relevant for industry purposes, however it is equally important to produce results in the shortest time possible, mainly for latency-sensitive applications and to control cloud hosting costs. In this paper we show the effectiveness of translating with 8-bit quantization for models that have been trained using 32-bit floating point values. Results show that 8-bit translation makes a non-negligible impact in terms of speed with no degradation in accuracy and adequacy.
\end{abstract}

\section{Introduction}

Neural machine translation (NMT) \cite{DBLP:journals/corr/BahdanauCB14,DBLP:conf/nips/SutskeverVL14} has recently achieved remarkable performance improving fluency and adequacy over phrase-based machine translation and is being deployed in commercial settings \cite{koehn-knowles:2017:NMT}.  However, this comes at a cost of slow decoding speeds compared to phrase-based and syntax-based SMT (see section \ref{eightdecoding}). 

NMT models are generally trained using 32-bit floating point
values.  At training time, multiple sentences can be processed in parallel leveraging graphical processing units (GPUs) to good advantage since the data is processed in batches.  This is also true for decoding for non-interactive applications such as bulk document translation.

Why is fast execution on CPUs important? First, CPUs are cheaper than GPUs. Fast CPU computation will reduce commercial deployment costs.  Second, 
for low-latency applications such as speech-to-speech translation \cite{DBLP:conf/eacl/NeubigCGL17}, it is important to translate individual sentences quickly enough so that users can have an application experience that responds seamlessly.  Translating individual sentences with NMT requires many memory bandwidth intensive matrix-vector or matrix-narrow matrix multiplications \cite{Abdelfattah:2016:KOL:2935754.2818311}.  In addition, the batch size is 1 and GPUs do not have a speed advantage over CPUs due to the lack of adequate parallel work (as evidenced by increasingly difficult batching scenarios in dynamic frameworks \cite{dynet}).

Others have successfully used low precision approximations to neural net models. \newcite{google-speed} explored 8-bit quantization for feed-forward neural nets for speech recognition. \newcite{D17-1300} explored 16-bit quantization for machine translation. In this paper we show the effectiveness of 8-bit decoding for models that have been trained using 32-bit floating point values. Results show that 8-bit decoding does not hurt the fluency or adequacy of the output, while producing results up to 4-6x times faster.  In addition, implementation is straightforward and we can use the models as is without altering training.

The paper is organized as follows: Section \ref{model} reviews the attentional model of translation to be sped up, Section \ref{eightdecoding} presents our 8-bit quantization in our implementation, Section \ref{accuracy} presents automatic measurements of speed and translation quality plus human evaluations, Section \ref{disc} discusses the results and some illustrative examples, Section \ref{relwork} describes prior work, and Section \ref{conclfuture} concludes the paper.

\section{The Attentional Model of Translation}
\label{model}
Our translation system implements the attentional model of translation \cite{DBLP:journals/corr/BahdanauCB14} consisting of an encoder-decoder network with an attention mechanism. 

The encoder uses a bidirectional GRU recurrent neural network \cite{DBLP:journals/corr/ChoMBB14} to encode a source sentence ${\bf{x}}=(x_1,...,x_l)$, where $x_i$ is the embedding vector for the $i$th word and $l$ is the sentence length. The encoded form is a sequence of hidden states ${\bf{h}} = (h_1, ..., h_l)$ where each $h_i$ is computed as follows

\begin{equation}
 h_i = 
 \begin{bmatrix}
 \overleftarrow{{h}_i} \\
 \overrightarrow{{h}_i}
\end{bmatrix}
=
\begin{bmatrix}
\overleftarrow{f}(x_i, \overleftarrow{h}_{i+1}) \\
\overrightarrow{f}(x_i, \overrightarrow{h}_{i-1})
\end{bmatrix},
\end{equation}

where $\overrightarrow{h_0} = \overleftarrow{h_0} = 0$.  Here $\overleftarrow{f}$ and $\overrightarrow{f}$ are GRU cells.

Given $\bf{h}$,  the  decoder  predicts  the
target  translation ${\bf y}$  by computing the output token sequence $(y_1,...y_m)$, where
$m$ is the length of the sequence.
At each time $t$, the probability of each token
$y_t$ from a target vocabulary is
\begin{equation}
p(y_t | {\bf h}, y_{t-1}..y_1) = g(s_t, y_{t-1}, H_t),
\end{equation}
where
$g$ is a two layer feed-forward network over the embedding of the
previous target word ($y_{t-1}$), the decoder hidden state ($s_t$), and the weighted sum of encoder states
${\bf h}$ ($H_t$), followed by a softmax to predict the
probability distribution over the output vocabulary.

We compute $s_t$ with a two layer GRU as
\begin{equation}
s'_t = r(s_{t-1}, y^*_{t-1})
\end{equation}
and
\begin{equation}
s_t  = q(s'_t,H_t),
\end{equation}
where $s'_t$ is an intermediate state and $s_0=\overleftarrow{h_0}$.  The two GRU units $r$ and $q$ together with the attention constitute
the conditional GRU layer of \newcite{sennrich-EtAl:2017:EACLDemo}.  $H_t$ is computed as
\begin{equation}
H_t =
\begin{bmatrix}
\sum^l_{i=1} (\alpha_{t,i} \cdot \overleftarrow{h}_i) \\
\sum^l_{i=1} (\alpha_{t,i} \cdot \overrightarrow{h}_i)
\end{bmatrix},
\end{equation}
where $\alpha_{t,i}$ are the elements of $\alpha_{t}$ which is the output vector of the attention model.
This is computed with a two layer feed-forward network
\begin{equation}
\alpha'_t = v(\textrm{tanh}(w(h_i) + u(s'_{t-1}))),
\end{equation}
where $w$ and $u$ are weight matrices, and $v$ is another matrix resulting in one real value per encoder state $h_i$.  $\alpha_t$ is then the softmax over $\alpha'_t$.

We train our model using a program written using the Theano framework \cite{Bastien-Theano-2012}.  Generally models are trained with batch sizes ranging from 64 to 128 and unbiased Adam stochastic optimizer \cite{DBLP:journals/corr/KingmaB14}.  We use an embedding size of 620 and hidden layer sizes of 1000.  We select model parameters according to the best BLEU score on a held-out development set over 10 epochs.

\section{8-bit Translation} \label{eightdecoding}

Our translation engine is a C++ implementation. The engine is implemented using the Eigen matrix library, which provides efficient matrix operations. Each CPU core translates a single sentence at a time.  The same engine supports both batch and interactive applications, the latter making single-sentence translation latency important.  We report speed numbers as both words per second (WPS) and words per core second (WPCS), which is WPS divided by the number of cores running.  This gives us a measure of overall scaling across many cores and memory buses as well as the single-sentence speed.


Phrase-based SMT systems, such as  \cite{Tillmann:2006:EDP:1631828.1631830}, for English-German run at 170 words per core second (3400 words per second) on a 20 core Xeon 2690v2 system.  Similarly, syntax-based SMT systems, such as \cite{Zhao:2008:GLN:1613715.1613785}, for the same language pair run at 21.5 words per core second (430 words per second).

In contrast, our NMT system (described in Section \ref{model}) with 32-bit decoding runs at 6.5 words per core second (131 words per second). Our goal is to increase decoding speed for the NMT system to what can be achieved with phrase-based systems while maintaining the levels of fluency and adequacy that NMT offers. 

Benchmarks of our NMT decoder unsurprisingly show matrix multiplication as the number one source of compute cycles. In Table \ref{preprofile} we see that more than 85\% of computation is spent in Eigen's matrix and vector multiply routines (Eigen matrix vector product and Eigen matrix multiply). It dwarfs the costs of the transcendental function computations as well as the bias additions.

\begin{table}[!ht]
\centering
\begin{tabular}{S[table-format=2.2,table-space-text-post=\%] | c}
{Time} & Function \\
\hline
50.77\% & Eigen matrix vector product \\
35.02\% & Eigen matrix multiply \\
1.95\%  & NMT decoder layer \\
1.68\%  & Eigen fast tanh \\
1.35\%  & NMT tanh wrapper \\
\end{tabular}
\caption{Profile before 8-bit conversion. More than 85\% is spent in Eigen matrix/vector multiply routines.}
\label{preprofile}
\end{table}

Given this distribution of computing time, it makes sense to try to accelerate the matrix operations as much as possible.
One approach to increasing speed is to quantize matrix operations.  Replacing 32-bit floating point math operations with 8-bit integer approximations in neural nets has been shown to give speedups and similar accuracy \cite{google-speed}.  We chose to apply similar optimization to our translation system, both to reduce memory traffic as well as increase parallelism in the CPU.

Our 8-bit matrix multiply routine uses a naive implementation with no blocking or copy.  The code is implemented using Intel SSE4 vector instructions and computes 4 rows at a time, similar to \cite{D17-1300}. Simplicity led to implementing 8-bit matrix multiplication with the results being placed into a 32-bit floating point result.  This has the advantage of not needing to know the scale of the result.  In addition, the output is a vector or narrow matrix, so little extra memory bandwidth is consumed.

Multilayer matrix multiply algorithms result in significantly faster performance than naive algorithms \cite{Goto:2008:AHM:1356052.1356053}.  This is due to the fact that there are $O(N^3)$ math operations on $O(N^2)$ elements when multiplying NxN matrices, therefore it is worth significant effort to minimize memory operations while maximizing math operations. However, when multiplying an NxN matrix by an NxP matrix where P is very small ($<$10), memory operations dominate and performance does not benefit from the complex algorithm.  When decoding single sentences, we typically set our beam size to a value less than 8 following standard practice in this kind of systems \cite{koehn-knowles:2017:NMT}.  We actually find that at such small values of P, the naive algorithm is a bit faster.

\begin{table}[!ht]
\centering
\begin{tabular}{S[table-format=2.2,table-space-text-post=\%] | c}
{Time} & Function \\
\hline
69.54\% & 8-bit matrix multiply \\
6.37\%  & Eigen fast tanh \\
2.06\%  & NMT decoder layer \\
0.95\%  & NMT tanh wrapper \\
\end{tabular}
\caption{Profile after 8-bit conversion.  Matrix multiply includes matrix-vector multiply. Matrix multiply is still 70\% of computation.  Tanh is larger but still relatively small.}
\label{postprofile}
\end{table}

Table \ref{postprofile} shows the profile after converting the matrix routines to 8-bit integer computation. There is only one entry for matrix-matrix and matrix-vector multiplies since they are handled by the same routine. After conversion, tanh and sigmoid still consume less than 7\% of CPU time.  We decided not to convert these operations to integer in light of that fact.

It is possible to replace all the operations with 8-bit approximations
\cite{DBLP:journals/corr/WuSCLNMKCGMKSJL16}, but this makes implementation more complex, as the scale of the result of a matrix multiplication must be known to correctly output 8-bit numbers without dangerous loss of precision.

Assuming we have 2 matrices of size 1000x1000 with a range of values $[-10,10]$, the individual dot products in the result could be as large as $10^8$.  In practice with neural nets, the scale of the result is similar to that of the input matrices.  So if we scale the result to $[-127,127]$ assuming the worst case, the loss of precision will give us a matrix full of zeros.  The choices are to either scale the result of the matrix multiplication with a reasonable value, or to store the result as floating point. We opted for the latter.

8-bit computation achieves 32.3 words per core second (646 words per second), compared to the 6.5 words per core second (131 words per second) of the 32-bit system (both systems load parameters from the same model). This is even faster than the syntax-based system that runs at 21.5 words per core second (430 words per second). Table \ref{finalspeeds} summarizes running speeds for the phrase-based SMT system, syntax-based system and NMT with 32-bit decoding and 8-bit decoding.

\begin{table}[!ht]
\centering
\begin{tabular}{c | S[table-format=3.1]}
System & WPCS \\
\hline
Phrase-based  & 170\\
Syntax-based  & 21.5 \\
NMT 32-bit  & 6.5 \\
NMT 8-bit  & 32.3 \\
\end{tabular}
\caption{Running speed (in words per core second) of the phrase-based SMT system, syntax-based system, NMT with 32-bit decoding and NMT with 8-bit decoding.}
\label{finalspeeds}
\end{table}

\section{Measurements}\label{accuracy}
To demonstrate the effectiveness of approximating the floating point math with 8-bit integer computation, we show automatic evaluation results on several models, as well as independent human evaluations.  We report results on Dutch-English, English-Dutch, Russian-English, German-English and English-German models.  Table \ref{datasize} shows training data sizes and vocabulary sizes.  All models have 620 dimension embeddings and 1000 dimension hidden states.

\begin{table}[!ht]
\centering
\begin{tabular}{c|c|c|c}
Lang & Training & Source & Target \\
 & Sentences & Vocabulary & Vocabulary \\
\hline
En-Nl & 17M & 42112 & 33658 \\
Nl-En & 17M & 33658 & 42212 \\
\hline
Ru-En & 31M & 42388 & 42840 \\
\hline
En-De & 31M & 57867 & 63644 \\
De-En & 31M & 63644 & 57867 \\
\end{tabular}
\caption{Model training data and vocabulary sizes}
\label{datasize}
\end{table}

\subsection{Automatic results}

Here we report automatic results comparing decoding results on 32-bit and 8-bit implementations.  As others have found \cite{DBLP:journals/corr/WuSCLNMKCGMKSJL16}, 8-bit implementations impact quality very little.

In Table \ref{results1}, we compared automatic scores and speeds for Dutch-English, English-Dutch, Russian-English, German-English and English-German models on news data. The English-German model was run with both a single model (1x) and an ensemble of two models (2x) \cite{DBLP:journals/corr/FreitagAS17}.  Table \ref{testsize} gives the number of sentences and average sentence length for the test sets used.

\begin{table}[!ht]
\centering
\begin{tabular}{c|c|c|c}
Lang & Test & Src Sent & Tgt Sent\\
 & Sentences & Length & Length \\
\hline
En-Nl & 990 & 22.5 & 25.9 \\
Nl-En & 990 & 25.9 & 22.5 \\
\hline
Ru-En & 555 & 27.2 & 35.2 \\
\hline
En-De & 168 & 51.8 & 46.0 \\
De-En & 168 & 46.0 & 51.8 \\
\end{tabular}
\caption{Test data sizes and sentence lengths}
\label{testsize}
\end{table}

\begin{table} [!ht]
\centering
\begin{tabular}{c|c| S[table-format=2.1]| S[table-format=2.1]}
Lang & Mode & {BLEU} & {Speed (WPSC)} \\
\hline
En-Nl & 32-bit &  31.2 & 12.6 \\
En-Nl & 8-bit &  31.2 & 58.9 \\
\hline
Nl-En & 32-bit &  36.1 & 10.3 \\
Nl-En & 8-bit &  36.3 & 45.8 \\
\hline
Ru-En & 32-bit &  24.5 & 8.9 \\
Ru-En & 8-bit &  24.3 & 51.4 \\
\hline
De-En & 32-bit &  32.6 & 7.3 \\
De-En & 8-bit &  32.2 & 37.5 \\
\hline
En-De 2x & 32-bit &  30.5 & 7.1 \\
En-De 2x & 8-bit &  30.6 & 33.7 \\
\hline
En-De 1x & 32-bit &  29.7 & 15.9 \\
En-De 1x & 8-bit &  29.7 & 71.3 \\
\end{tabular}
\caption{BLEU scores and speeds for 8-bit and 32-bit versions of several models.  Speeds are reported in words per core second.}
\label{results1}
\end{table}

Speed is reported in words per core second (WPCS).  This gives us a better sense of the speed of individual engines when deployed on multicore systems with all cores performing translations. Total throughput is simply the product of WPCS and the number of cores in the machine. The reported speed is the median of 9 runs to ensure consistent numbers.  The results show that we see a 4-6x speedup over 32-bit floating point decoding.  German-English shows the largest deficit for the 8-bit mode versus the 32-bit mode.  The German-English test set only includes 168 sentences so this may be a spurious difference.

\subsection{Human evaluation}

These automatic results suggest that 8-bit quantization can be done without
perceptible degradation.  To confirm this, we carried out a  human evaluation experiment.

In Table \ref{human}, we show the results of performing human evaluations on some of the same language pairs in the previous section. An independent native speaker of the language being translated to/from different than English (who is also proficient in English) scored 100 randomly selected sentences. The sentences were shuffled during the evaluation to avoid evaluator bias towards different runs. We employ a scale from 0 to 5, with 0 being unintelligible and 5 being perfect translation. 
\begin{table}[!ht]
\centering
\begin{tabular}{c|c|c}
Language & 32-bit & 8-bit \\
\hline
En-Nl & 4.02 & 4.08 \\
Nl-En & 4.03 & 4.03 \\
\hline
Ru-En & 4.10 & 4.06 \\
\hline
En-De 2x & 4.05 & 4.16 \\
En-De 1x & 3.84 & 3.90 \\
\end{tabular}
\caption{Human evaluation scores for 8-bit and 32-bit systems. All tests are news domain.}
\label{human}
\end{table}

\begin{table*}[!ht]
\centering
\begin{tabular}{c|c|l}
Source & Time & Sie standen seit 1946 an der Parteispitze \\
32-bit & 720 ms & They \textbf{had been} at the \textbf{party leadership} since 1946 \\
8-bit & 180 ms & They \textbf{stood} at the \textbf{top of the party} since 1946. \\
\hline
Source & Time & So erwarten die Experten f\"ur dieses Jahr lediglich einen Anstieg der Weltproduktion \\&& von 3,7 statt der im Juni prognostizierten 3,9 Prozent. F\"ur 2009 sagt das  Kieler \\&& Institut sogar eine Abschw\"achung auf 3,3 statt 3,7 Prozent voraus.  \\
32-bit & 4440 ms & For this year, the experts expect only an increase in world production of 3.7 \\&& instead of the 3.9 percent forecast in June. In 2009, the Kiel Institute \\&& \textbf{predictated} a slowdown to \textbf{3.3 percent} instead of 3.7 \textbf{percent}. \\
8-bit & 750 ms & For this year, the experts expect only an increase in world production of 3.7 \\&& instead of the 3.9 percent forecast in June. In 2009, the Kiel Institute \textbf{even} \\&& \textbf{forecast} a slowdown to \textbf{3.3\%} instead of 3.7 \textbf{per cent}.  \\
\hline
Source & Time & Heftige Regenf\"alle wegen ``Ike" werden m\"oglicherweise schwerere Sch\"aden anrichten \\&& als seine Windb\"oen. Besonders gef\"ahrdet sind dicht besiedelte Gebiete im Tal des Rio \\&& Grande, die noch immer unter den Folgen des Hurrikans ``Dolly" im Juli leiden. \\
32-bit & 6150 ms & Heavy rainfall due to ``Ike" may cause \textbf{more severe} damage than its gusts of wind, \\&& particularly in densely populated areas in the Rio Grande valley, which \textbf{are still} \\&& \textbf{suffering} from the consequences of the ``dolly" hurricane in July. \\
8-bit & 1050 ms & Heavy rainfall due to ``Ike" may cause \textbf{heavier} damage than its gusts of wind, \\&& particularly in densely populated areas in the Rio Grande valley, which \textbf{still} \\&& \textbf{suffer} from the consequences of the ``dolly" hurricane in July. \\
\end{tabular}
\caption{Examples of De-En news translation system comparing 32-bit and 8-bit decoding.  Differences are in boldface.  Sentence times are average of 10 runs. }
\label{examples}
\end{table*}

\begin{table*}[!ht]
\centering
\begin{tabular}{c|c|l}
Source & Time & Het is tijd om de kloof te overbruggen. \\
32-bit & 730 ms & \textbf{It's} time to bridge the gap. \\
8-bit & 180 ms& \textbf{It is} time to bridge the gap. \\
\hline
Source & Time & Niet dat Barientos met zijn vader van plaats zou willen wisselen. \\
32-bit &  1120 ms & Not that Barientos would \textbf{want} to \textbf{change his father's place}. \\
8-bit & 290 ms & Not that Barientos would \textbf{like} to \textbf{switch places with his father}. \\
\end{tabular}
\caption{Examples of Nl-En news translation system comparing 32-bit and 8-bit decoding.  Differences are in boldface. Sentence times are average of 10 runs. }
\label{examples2}
\end{table*}

The Table shows that the automatic scores shown in the previous section are also sustained by humans. 8-bit decoding is as good as 32-bit decoding according to the human evaluators. 

%
%
%
%
%
%
%
%
%
%
%
%
%
%

\section{Discussion}
\label{disc}

Having a faster NMT engine with no loss of accuracy is commercially useful.  In our deployment scenarios, it is the difference between an interactive user experience that is sluggish and one that is not.  Even in batch mode operation, the same throughput can be delivered with 1/4 the hardware.

In addition, this speedup makes it practical to deploy small ensembles of models.  As shown above in the En-De model in Table \ref{results1}, an ensemble can deliver higher accuracy at the cost of a 2x slowdown.  This work makes it possible to translate with higher quality while still being at least twice as fast as the previous baseline.

As the numbers reported in Section \ref{accuracy} demonstrate, 8-bit and 32-bit decoding have similar average quality.  As expected, the outputs produced by the two decoders are not identical.  In fact, on a run of 166 sentences of De-En translation, only 51 were identical between the two. In addition, our human evaluation results and the automatic scoring suggest that there is no specific degradation by the 8-bit decoder compared to the 32-bit decoder. In order to emphasize these claims, Table \ref{examples} shows several examples of output from the two systems for a German-English system.  Table \ref{examples2} shows 2 more examples from a Dutch-English system.

In general, there are minor differences without any loss in adequacy or fluency due to 8-bit decoding.  Sentence 2 in Table \ref{examples} shows a spelling error (``predictated") in the 32-bit output due to reassembly of incorrect subword units.\footnote{In order to limit the vocabulary, we use BPE subword units \cite{sennrich-haddow-birch:2016:P16-12} in all models.}

\section{Related Work}
\label{relwork}
Reducing the resources required for decoding neural nets in general and neural machine translation in particular has been the focus of some attention in recent years.  

\newcite{google-speed} explored accelerating convolutional neural nets with 8-bit integer decoding for speech recognition.  They demonstrated that low precision computation could be used with no significant loss of accuracy.
\citet{DBLP:journals/corr/HanMD15} investigated highly compressing image classification neural networks using network pruning, quantization, and Huffman coding so as to fit completely into on-chip cache, seeing significant improvements in speed and energy efficiency while keeping accuracy losses small.

Focusing on machine translation, \citet{D17-1300} implemented 16-bit fixed-point integer math to speed up
matrix multiplication operations, seeing a 2.59x improvement.  They show competitive BLEU scores on WMT English-French NewsTest2014 while offering significant speedup.
Similarly, \cite{DBLP:journals/corr/WuSCLNMKCGMKSJL16} applies 8-bit end-to-end quantization in translation models.  They also show that automatic metrics do not suffer as a result.  In this work, quantization requires modification to model training to limit the size of matrix outputs.

\section{Conclusions and Future Work}
\label{conclfuture}
In this paper, we show that 8-bit decoding for neural machine translation runs up to 4-6x times faster than a similar optimized floating point implementation.  We show that the quality of this approximation is similar to that of the 32-bit version.  We also show that it is unnecessary to modify the training procedure to produce models compatible with 8-bit decoding. 

To conclude, this paper shows that 8-bit decoding is as good as  32-bit decoding both in automatic measures and from a human perception perspective, while it improves latency substantially.

In the future we plan to implement a multi-layered matrix multiplication that falls back to the naive algorithm for matrix-panel multiplications.  This will provide speed for batch decoding for applications that can take advantage of it. We also plan to explore training with low precision for faster experiment turnaround time.

Our results offer hints of improved accuracy rather than just parity.  Other work has used training as part of the compression process.  We would like to see if training quantized models changes the results for better or worse.

\bibliography{naaclhlt2018}
\bibliographystyle{acl_natbib}

\end{document}